\documentclass{article}

\usepackage[preprint]{neurips_2023}
\usepackage{amssymb}
\usepackage{epstopdf}
\usepackage{overpic}
\usepackage{bm}
\usepackage{multirow}
\usepackage{color}
\usepackage{multirow}
\usepackage{algorithm}
\usepackage{algorithmic}
\usepackage{multicol}
\usepackage{siunitx}
\usepackage{amsthm}
\usepackage{makecell}
\usepackage{bbding}
\usepackage{balance}
\usepackage{lipsum}

\usepackage[utf8]{inputenc} % allow utf-8 input
\usepackage[T1]{fontenc}    % use 8-bit T1 fonts
\usepackage{hyperref}       % hyperlinks
\usepackage{url}            % simple URL typesetting
\usepackage{booktabs}       % professional-quality tables
\usepackage{amsfonts}       % blackboard math symbols
\usepackage{nicefrac}       % compact symbols for 1/2, etc.
\usepackage{microtype}      % microtypography
\usepackage{xcolor}         % colors
\usepackage{natbib}
\setcitestyle{numbers,square}

\title{CausalVLR: A Toolbox and Benchmark for Visual-Linguistic Causal Reasoning}

\author{%
  Yang Liu\\
Sun Yat-sen University\\
  \texttt{liuy856@mail.sysu.edu.cn} \\
  \And
  Weixing Chen \\
 Sun Yat-sen University\\
  \texttt{chen867820261@gmail.com} \\
    \AND
 Guanbin Li \\
 Sun Yat-sen University\\
\texttt{liguanbin@mail.sysu.edu.cn} \\
  \AND
 Liang Lin \\
 Sun Yat-sen University\\
  \texttt{linliang@ieee.org} \\
}

\begin{document}

\maketitle

\begin{abstract}
We present CausalVLR (Causal Visual-Linguistic Reasoning), an open-source toolbox containing a rich set of state-of-the-art causal relation discovery and causal inference  methods for various visual-linguistic reasoning tasks, such as VQA, image/video captioning, medical report generation, model generalization and robustness, etc. These methods have been included in the toolbox with PyTorch implementations under NVIDIA  computing system.
It not only includes training and inference codes, but also provides model weights. We believe
this toolbox is by far the most complete visual-linguitic causal reasoning toolbox. We wish that the toolbox and benchmark could serve the growing research community by providing a flexible toolkit
to re-implement existing methods and develop their own new causal reasoning methods. Code and models are available at \url{https://github.com/HCPLab-SYSU/CausalVLR}. The project is under active development by HCP-Lab\footnote{https://www.sysu-hcp.net}'s contributors and we will keep this document updated.
\end{abstract}

\section{Introduction}

The emergence of vast amounts of heterogeneous multi-modal data, including images \cite{he2016deep,liu2016combining}, videos \cite{liu2018transferable,liu2018global,liu2018hierarchically,liu2022tcgl,yan2023skeletonmae}, languages \cite{CMCIR,liu2023causality,wei2023visual}, audios \cite{lan2022audio}, and multi-sensor \cite{liu2019deep,liu2021semantics,zhu2022hybrid,ni2022cross,wang2023urban,lin2023denselight} data, has led to the application of large language models (LLMs) such as ChatGPT \cite{van2023chatgpt} and ChatGLM \cite{zeng2022glm} in various vision and language 
tasks, showing promising performance. However, current LLMs heavily rely on fitting extensive knowledge distributions, often capturing spurious correlations across different modalities. Consequently, they struggle to learn reliable chain-of-thought (COT) \cite{weichain} that reflects essential causal relations within multi-modal knowledge, limiting their generalization and cognitive abilities. Fortunately, causal inference \cite{pearl2009causality,scholkopf2021toward,liu2022causal,liu2022causalarxiv} ] provides a promising alternative for learning robust and reliable cross-modal models due to its promising ability to achieve robust representation and model learning with good cognitive ability. For a detailed review of causal inference and visual representation learning, please refer to our review paper \cite{liu2022causal}.

\begin{figure}[!t]
\centerline{\includegraphics[scale=0.43]{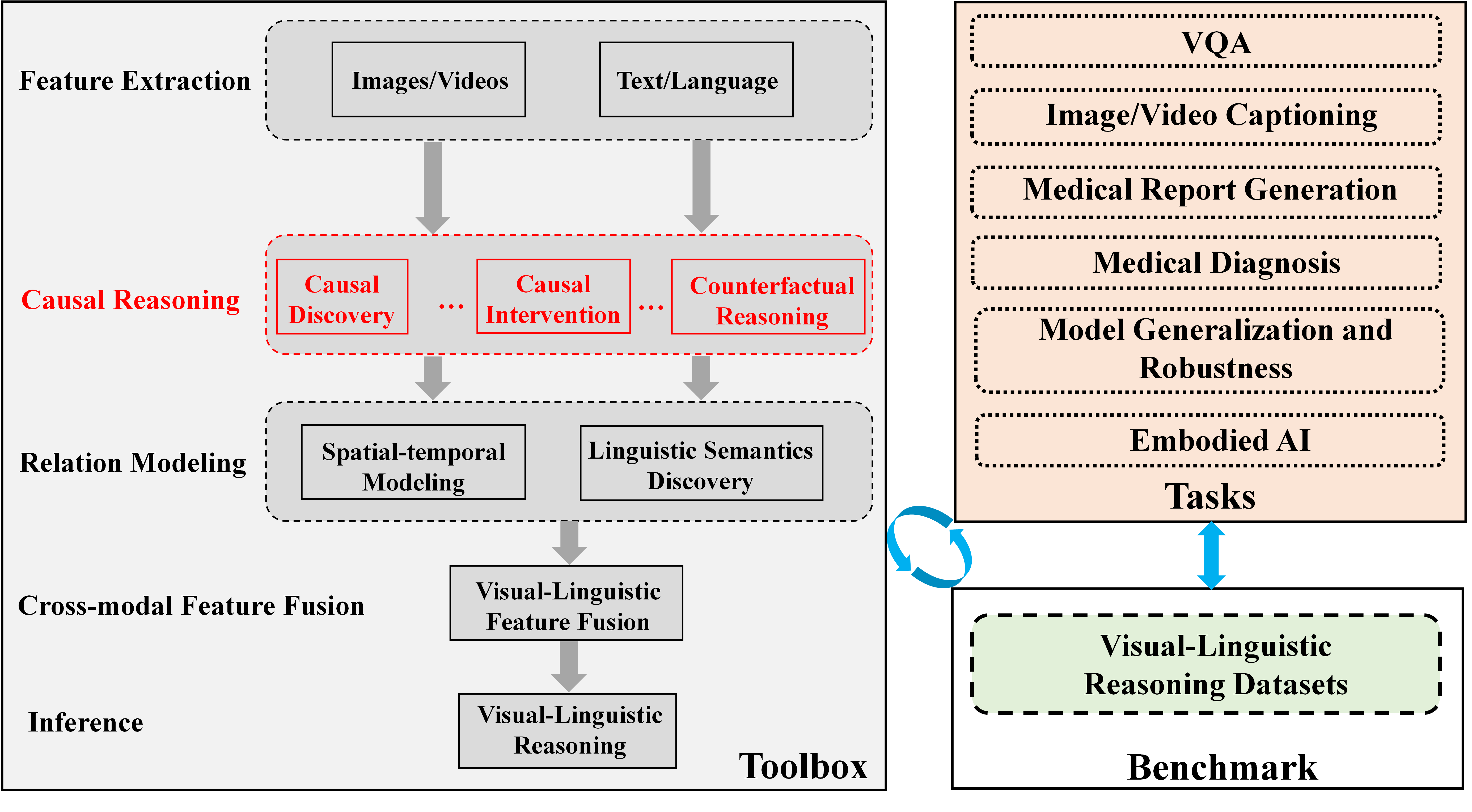}}
\caption{Overview of CausalVLR. CausalVLR is a python open-source framework for causal relation discovery, causal inference that implements state-of-the-art causal learning algorithms for various visual-linguistic reasoning tasks, such as VQA, image/video captioning,  medical report generation, medical diagnosis, model generalization and robustness, etc.}
\label{fig1}
\end{figure}

Visual-linguistic reasoning endeavors to comprehend both visual and linguistic content while performing various reasoning tasks, such as visual question answering (VQA), visual dialog, image/video captioning, and medical report generation. However, to date, there has been no comprehensive open-source framework available for causality-aware visual-linguistic reasoning. With the aim of offering a high-quality toolbox and a unified benchmark, we have developed CausalVLR: a pytorch-based open-source toolbox and benchmark designed specifically for visual-linguistic causal reasoning. Figure \ref{fig1} provides an overview of CausalVLR.

CausalVLR offers several key features: (1) Modular design: We decompose the visual-linguistic reasoning framework into different components, allowing for the easy construction of customized visual-linguistic reasoning frameworks by combining various modules. (2) Support for multiple frameworks out of the box: The toolbox provides support for popular and contemporary visual-linguistic reasoning frameworks. (3) High efficiency: All basic modules and operations are executed on GPUs to ensure optimal performance. (4) State of the art: The toolbox is derived from the codebase developed by the experienced HCP-Lab team, specializing in causal inference and visual-linguistic reasoning, and continuous improvements are made. In addition to introducing the codebase and benchmark results, we also share our experiences and best practices for visual-linguistic causal reasoning. Ablation experiments involving hyperparameters, architectures, and training strategies are conducted and discussed. Our aim is to contribute to future research and facilitate comparisons between different methods. The remaining sections are organized as follows: First, we present the various supported methods and highlight important features of CausalVLR. Next, we present the benchmark results. Finally, we showcase representative studies on selected baselines. Table \ref{tab1} provides a summary of representative algorithms from the HCP-Lab team.

\section{Algorithms}
This section provides a summary of three representative state-of-the-art (SOTA) algorithms for visual question answering (VQA) and medical report generation tasks. All algorithms have been implemented using PyTorch. The CausalVLR library will be continuously updated in the coming years. In this section, we will provide a concise introduction to the selected algorithms.

\begin{table}[!t]
  \centering
  %\begin{threeparttable}[c]
  \caption{Representative visual-linguistic causal reasoning algorithms in CausalVLR.}
  \small
  \setlength{\tabcolsep}{1pt}\renewcommand\arraystretch{2}
  \begin{tabular}{cccc}
  \hline
  \centering
  %\begin{tabular}{llllll}
    %\toprule
    Task  &Algorithm & Highlight& Pytorch   \\
    %\midrule
    \hline
     Causal CoT&CausalGPT \cite{tang2023towards}&Causality-aware CoT&Yes\\\hline
   VQA&CMCIR \cite{CMCIR,liu2022cross}&Causal front-door and back-door interventions &Yes\\
      VQA&VCSR \cite{wei2023visual}&Visual causal scene discovery&Yes\\\hline
      Image Captioning&AAAI2023 \cite{WuYangAAAI}& Knowledge consensus&Yes\\\hline
      Medical Report Generation&VLCI \cite{chen2023visual}&Implicit visual causal intervention&Yes\\\hline
      Medical Diagnosis&TPAMI2023 \cite{Junfan}&Causality-aware medical diagnosis&Yes\\\hline
      Model Generalization and Robustness&CVPR2023 \cite{xiao2023masked}&Counterfactual Fine-tuning&Yes\\
    %\bottomrule
    \hline
  \end{tabular}
    \vspace{-10pt}
  \label{tab1}
    %\end{threeparttable}
\end{table}

\subsection{CausalGPT}

\textbf{Caco-CoT} \cite{tang2023towards}. The framework of causal-consistency CoT (CaCo-CoT is shown in Figure \ref{fig:case}.The contribution of Caco-CoT is listed as follows:

\begin{figure*}[!t]
% \vspace{-10pt}
\center
\includegraphics[width=\textwidth]{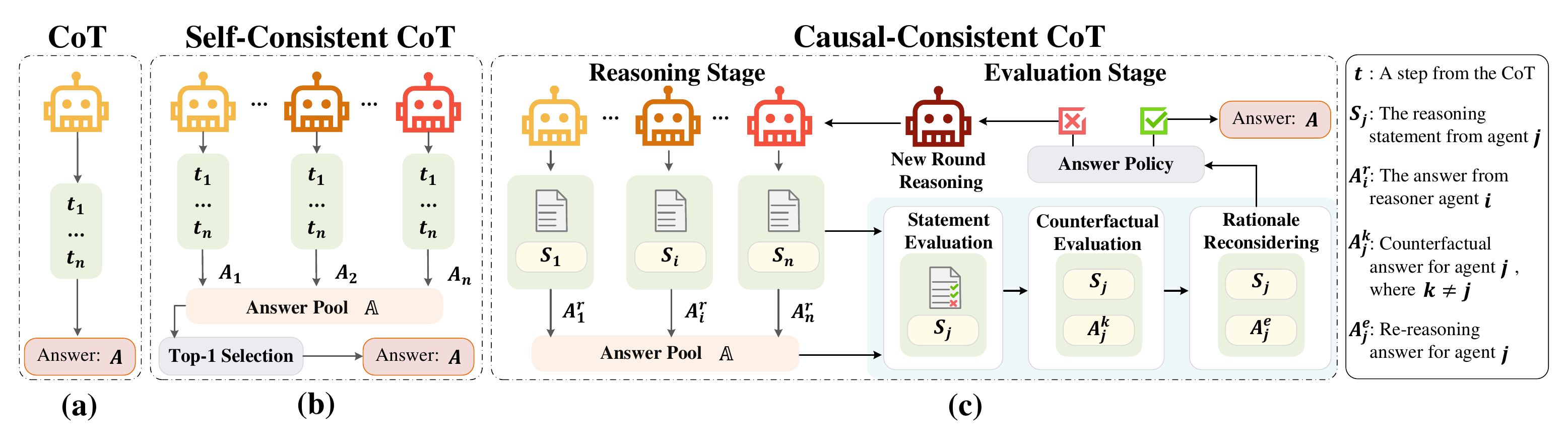}
\vspace{-20pt}
\caption{Comparison between existing approaches and causal-consistency CoT (CaCo-CoT). In CaCo-CoT, \textit{reasoner} agents and an \textit{evaluator} agent cooperate to facilitate a causally consistent reasoning process, thereby minimizing factual and inferential errors. } 
\label{fig:case}
\vspace{-5pt}
\end{figure*}

To ensure reasoning faithfulness of the solutions by inspecting causal consistency, an evaluator is introduced. Initially, the evaluator is prompted to examine the statements in the solution step by step while considering the whole reasoning structure, which is in a non-causal direction. For further causal consistency assessment, the evaluator then moves on
to apply a counterfactual answer to the original question
and look for any contradiction therein. Finally, it turns out a modified answer after its reconsideration. The group of agents cooperates for a consensual answer if the most frequent solution is accepted by the evaluator. In the tasks of science question answering and commonsense reasoning, we demonstrate that our causal-consistency framework outperforms the existing approaches through a series of experiments and comparisons.

The official codes and data are available at \url{https://github.com/HCPLab-SYSU/CausalVLR}.

\subsection{VQA}

\textbf{CMCIR}\cite{CMCIR}. The contribution of CMCIR is listed as follows:
\begin{itemize}
\item We propose a causality-aware event-level visual question answering framework named Cross-Modal Causal RelatIonal Reasoning (CMCIR), to discover true causal structures via causal intervention on the integration of visual and linguistic modalities and
achieve robust event-level visual question answering performance. To the best of our knowledge, we are the first to discover cross-modal causal structures for the event-level visual question answering task.

\item  We introduce a linguistic back-door causal intervention module guided by linguistic semantic relations
to mitigate the spurious biases and uncover the causal dependencies within the linguistic modality. To disentangle the visual spurious correlations, we
propose a Local-Global Causal Attention Module (LGCAM) that aggregates the local and global visual representations by front-door causal intervention.

\item We construct a Spatial-Temporal Transformer (STT) that models the multi-modal co-occurrence interactions between the visual and linguistic knowledge, to discover the fine-grained interactions among linguistic semantics, spatial, and temporal representations.

\item To adaptively fuse the causality-aware visual and
linguistic features, we introduce a Visual-Linguistic
Feature Fusion (VLFF) module that leverages the
hierarchical linguistic semantic relations to learn the
global semantic-aware visual-linguistic features.

\item Extensive experiments on SUTD-TrafficQA, TGIF-QA, MSVD-QA, and MSRVTT-QA datasets show the effectiveness of our CMCIR for discovering visuallinguistic causal structures and achieving promising
event-level visual question answering performance.
\end{itemize}

The official codes and data are available at \url{https://github.com/HCPLab-SYSU/CMCIR}.

\textbf{VCSR}\cite{wei2023visual}. The contribution of CMCIR is listed as follows:
\begin{itemize}
\item We propose the Visual Causal Scene Refinement (VCSR) framework, to explicitly discover true causal
visual scenes from the perspective of causal front-door intervention. To the best of our knowledge, we are the first to discover visual causal scenes for video question answering.
\item We build the Causal Scene Separator (CSS) module that learns to discover a collection of visual causal and non-causal scenes based on the visual-linguistic causal relevance and estimates the causal effect of the scene-separating intervention in a contrastive learning manner.
\item We introduce the Question-Guided Refiner (QGR) module that refines consecutive video frames guided
by the question semantics to obtain more representative segment features for causal front-door intervention.
\end{itemize}

The official codes and data are available at \url{https://github.com/HCPLab-SYSU/CausalVLR}.

\subsection{Medical Report Generation}

\textbf{VLCI}\cite{chen2023visual}. The contribution of CMCIR is listed as follows:
\begin{itemize}
\item To implicitly mitigate cross-modal confounders and
discover the true cross-modal causality, we propose
visual-linguistic causal front-door intervention modules VDM and LDM. The VDM aims to disentangle
the region-based features from images in the encoder,
and the LDM aims to eliminate the spurious correlations caused by the visual-linguistic embedding.

\item To alleviate the problem of unpaired data when pretraining visual-linguistic RRG data, we combine the
PLM and MIM for cross-modal pre-training in various
data situations (e.g., unpaired, single modality), which
is efficient and easy to implement.

\item We propose a lightweight Visual-Linguistic Causal Intervention (VLCI) framework for RRG, which introduces mediators without additional knowledge, to implicitly deconfound the visual-linguistic confounder
by causal front-door intervention. Experimental results show that VLCI achieves state-of-the-art performance on two datasets IU-Xray and MIMIC-CXR.
\end{itemize}
The official codes and data are available at \url{https://github.com/WissingChen/VLCI}.

\section{Conclusions and future works}
This paper presents CausalVLR, an open-source toolbox that offers a comprehensive collection of state-of-the-art methods for causal relation discovery and causal inference in visual-linguistic reasoning tasks. We highlight the contributions of representative state-of-the-art visual-linguistic causal reasoning methods. CausalVLR aims to enhance the development of causal inference in the visual-linguistic reasoning domain by providing easily accessible open-source implementations, benchmarking models, and computational resources. We plan to incorporate additional state-of-the-art algorithms and benchmarks in future updates.

\bibliographystyle{plain}
\bibliography{refs}

%%%%%%%%%%%%%%%%%%%%%%%%%%%%%%%%%%%%%%%%%%%%%%%%%%%%%%%%%%%%

\end{document}